\PassOptionsToPackage{hyphens}{url}
\documentclass{article}
\usepackage{spconf,amsmath,graphicx,hyperref,amsfonts}
\makeatletter
\def\url@leostyle{%
  \@ifundefined{selectfont}{\def\UrlFont{\sf}}{\def\UrlFont{\small\ttfamily}}}
\makeatother
\newlength{\bibitemsep}
\newlength{\bibparskip}
\let\oldthebibliography\thebibliography
\renewcommand\thebibliography[1]{%
  \oldthebibliography{#1}%
  \setlength{\parskip}{\bibitemsep}%
  \setlength{\itemsep}{\bibparskip}%
}

\def\RR{\mathbb R}
\def\thline{\noalign{\hrule height 1.0pt}}

\title{Deep Clustering and Conventional Networks for Music Separation: Stronger Together}
\name{Yi Luo$^{\dagger}$ \qquad Zhuo Chen$^{\dagger}$ \qquad John R. Hershey$^{\ddagger}$ \qquad Jonathan Le Roux$^{\ddagger}$ \qquad Nima Mesgarani$^{\dagger}$}

\address{$^{\dagger}$ Department of Electrical Engineering, Columbia University, New York, NY\\
    $^{\ddagger}$Mitsubishi Electric Research Laboratories (MERL), Cambridge, MA}
\begin{document}
\ninept
\maketitle
\begin{abstract}
Deep clustering is the first method to handle general audio separation scenarios with multiple sources of the same type and an arbitrary number of sources,  performing impressively in speaker-independent speech separation tasks. However, little is known about its effectiveness in other challenging situations such as music source separation. 
Contrary to conventional networks that directly estimate the source signals, deep clustering generates an embedding for each time-frequency bin, and separates sources by clustering the bins in the embedding space.
We show that deep clustering outperforms conventional networks on a singing voice separation task, in both matched and mismatched conditions, even though conventional networks have the advantage of end-to-end training for best signal approximation, presumably because its more flexible objective engenders better regularization. 
Since the strengths of deep clustering and conventional network architectures appear complementary, we explore combining them in a single hybrid network trained via an approach akin to multi-task learning. Remarkably, the combination significantly outperforms either of its components. 
\end{abstract}
\begin{keywords}
Deep clustering, Singing voice separation, Music separation, Deep learning
\end{keywords}
\section{Introduction}
\label{sec:intro}

Monaural music source separation has been the focus of many research efforts for over a decade.
This task aims at separating a music recording into several tracks where each track corresponds to a single instrument. 
A related goal is to design
algorithms that can separate vocals and accompaniment, where all the instruments are considered as one source. Music source separation algorithms have been successfully used for predominant pitch tracking \cite{fan2016singing}, accompaniment generation for Karaoke systems \cite{tachibana2016real}, or singer identification~\cite{berenzweig2002using}. 

Despite these advances, a system that can successfully generalize to different music datasets has thus far remained unachievable, due to the tremendous variability of music recordings, for example in terms of genre or types of instruments used. 
Unsupervised methods, such as those based on computational auditory scene analysis (CASA) \cite{li2007separation}, source/filter modeling \cite{durrieu2010source}, or low-rank and sparse modeling \cite{huang2012singing}, have difficulty in capturing the dynamics of the vocals and instruments, while supervised methods, such as those based on non-negative matrix factorization (NMF) \cite{sprechmann2012real}, F0-based estimation \cite{hsu2010improvement}, or Bayesian modeling \cite{yang2014bayesian}, suffer from generalization and processing speed issues.





Recently, deep learning has found many successful applications in audio source separation. Conventional regression-based networks try to infer the source signals directly, often by inferring time-frequency (T-F) masks to be applied to the T-F representation of the mixture so as to recover the original sources. These mask-inference networks have been shown to produce superior results compared to the traditional approaches in singing voice separation \cite{huang2014singing}. 
These networks are a natural choice when the sources can be characterized as belonging to distinct classes. 

Another promising approach designed for more general situations is the so-called deep clustering framework \cite{hershey2016deep}. Deep clustering has been applied very successfully to the task of single-channel speaker-independent speech separation \cite{hershey2016deep}. Because it uses of pair-wise affinities as separation criterion, deep clustering can handle mixtures with multiple sources from the same type, and an arbitrary number of sources. 
Such difficult conditions are endemic to music separation. 

In this study, we explore the use of both deep clustering and conventional mask-inference networks to separate the singing voice from the accompaniment, grouping all the instruments as one source and the vocals as another. The singing voice separation task that we consider here is amenable to class based separation, and would not seem to require the extra flexibility in terms of source types and number of sources that deep clustering would provide. However, in addition to opening up the potential to apply to more general settings, the additional flexibility of deep clustering may have some benefits in terms of regularization.   Whereas conventional mask-inference approaches only focus on increasing the separation between sources, the deep clustering objective also reduces within-source variance in the internal representation, which could be beneficial for generalization.  In recent work it has been shown that forcing deep network activations to cluster well can improve the resulting test performance \cite{liao2016learning}.  
To investigate these potential benefits, we develop a two-headed ``Chimera'' network with both a deep clustering head and a mask-inference head attached to the same network body. Each head has its own objective, but the whole hybrid network is trained in a joint fashion akin to multi-task training.
Our findings show that the addition of the deep clustering criterion greatly improves upon the performance of the mask-inference network. 



\section{Model Description}
\label{sec:format}
\subsection{Deep clustering}

Deep clustering operates according to the assumption that the T-F representation of the mixed signal can be partitioned into multiple sets, depending on which source is dominant (i.e., its power is the largest among all sources) in a given bin. A deep clustering
network takes features of the acoustic signal as input, and assigns a $D$-dimensional embedding  to  each T-F bin.  The network is trained to encourage the embeddings of T-F bins dominated by the same source to be similar to each other, and the embeddings of T-F bins dominated by different sources to be different. Note that the concept of ``source'' shall be defined according to the task at hand: for example, one speaker per source for speaker separation, all vocals in one source versus all other instruments in another source for singing voice separation, etc.
A T-F mask for separating each source can then be estimated by clustering the T-F embeddings \cite{isik2016single}. 

The training target is derived from a label indicator matrix ${\bf Y} \in \RR^{T  F \times C}$, where $T$ denotes the number of frames, $F$ the feature dimension, and $C$ the number of sources in the input mixture ${\bf x}$, such that $Y_{i,j} = 1$ if T-F  bin $i=(t,f)$ is dominated by source $j$, and $Y_{i,j} = 0$ otherwise. We can construct a binary affinity matrix ${\bf A} = {\bf Y}{\bf Y}^T$, which represents the assignment of the sources in a permutation independent way: $A_{i,j} = 1$ if $i$ and $j$ are dominated by the same source, and $A_{i,j} = 0$ if they are not. The network estimates an embedding matrix ${\bf V} \in \RR^{T F \times D}$, 
where $D$ is the embedding dimension. The corresponding estimated affinity matrix is then defined as ${\bf \hat{A}} = {\bf V}{\bf V}^T$. The cost function for the network is
\begin{equation}
    \mathcal{L}_{\text{DC}} = ||{\bf \hat{A}} - {\bf A}||_F^2 = ||{\bf V}{\bf V}^T - {\bf Y}{\bf Y}^T||_F^2. \label{eq:dc}
\end{equation}
Although the matrices ${\bf A}$ and ${\bf  \hat{A}}$ are typically very large, their low-rank structure can be exploited to decrease the computational complexity \cite{hershey2016deep}.

At test time,  a clustering algorithm such as K-means is applied to the embeddings $V$ to generate a cluster assignment matrix, which is used as a binary T-F mask applied to the mixture to estimate the T-F representation of each source.

\subsection{Multi-task learning and Chimera networks} 

Whereas the deep clustering objective function has been shown to enable the training of neural networks for challenging source separation problems, a disadvantage of deep clustering is that the post-clustering process needed to generate the mask and recover the sources is not part of the original objective function. 
On the other hand, for mask-inference networks, the objective function minimized during training is directly related to the signal recovery quality. 
We seek to combine the benefits of both approaches in a strategy reminiscent of  multi-task learning, except that here both approaches address the same separation task. 



In \cite{hershey2016deep} and \cite{isik2016single}, the typical structure of a deep clustering network is to have multiple stacked recurrent layers (e.g., BLSTMs) yielding an $N$-dimensional vector at the top layer, followed by a fully-connected linear layer. For each frame $t$, this layer outputs a $D$-dimensional vector for each of the $F$ frequencies, resulting in a $F \times D$ representation ${\bf Z}_t$. To form the embeddings, ${\bf Z}$ then passes through a $\mathrm{tanh}$ non-linearity, and unit-length normalization independently for each T-F bin. Concatenating across time results in the $TF \times D$ embedding matrix ${\bf V}$ as used in Eq.~\ref{eq:dc}.

\begin{figure}[t]
    \centering
    \includegraphics[width=7.5cm]{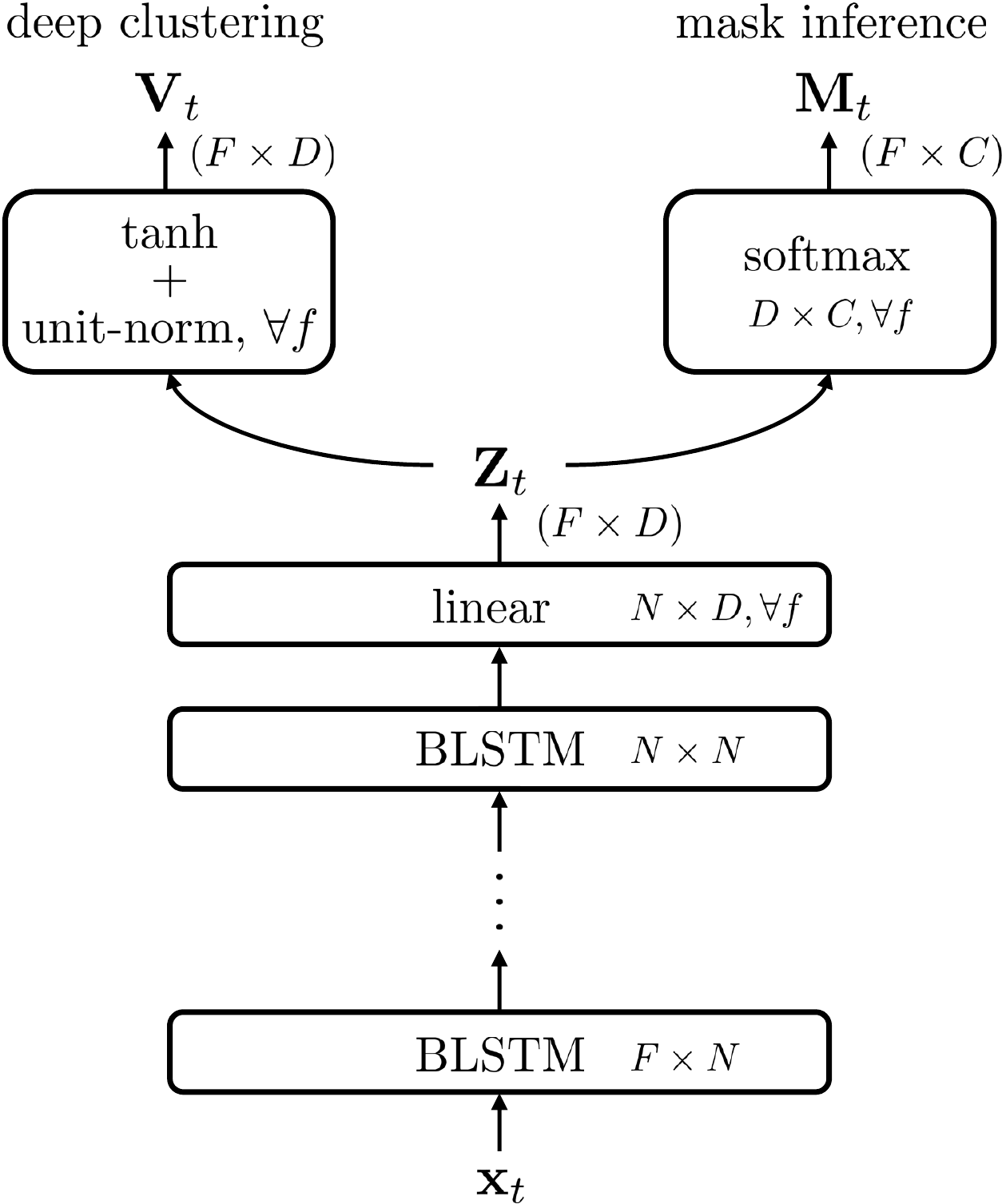}
    \vspace{-.2cm}
    \caption{Structure of the Chimera network.}
    \vspace{-.3cm}
    \label{fig:network_structure}
\end{figure}

We extend this architecture in order to create a two-headed network, which we refer as ``Chimera'' network, with one head outputting embeddings as in a deep clustering network, and the other head outputting a soft mask, as in a mask-inference network.
The new mask-inference head is obtained starting with ${\bf Z}$, and passing it through $F$  fully-connected $D \times C$ mask estimation layers (e.g., softmax), one for each frequency, resulting in $C$ masks ${\bf M}^{(c)}$, one for each source. 
The structure of the Chimera network is illustrated in Figure~\ref{fig:network_structure}.

The body of the network, up to the layer outputting ${\bf Z}$, can be trained with each head separately. For the deep clustering head, we use the objective $\mathcal{L}_{\text{DC}}$. 
For the mask-inference head, we can use a classical magnitude spectrum approximation (MSA) objective \cite{huang2012singing,Weninger2014GlobalSIP12,erdogan2015phase}, defined as:
\begin{equation}
    \mathcal{L}_{\text{MSA}} 
    = \sum_c ||{\bf R}^{(c)} - {\bf M}^{(c)} \odot {\bf S}||_2^2,
\end{equation}
where ${\bf R}^{(c)}$ denotes the magnitude of the T-F representation for the $c$-th clean source and ${\bf S}$ that of the mixture. 
Although this objective function makes sense intuitively, one caveat is that the mixture magnitude ${\bf S}$ may be smaller than that of a given source ${\bf R}^{(c)}$ due to destructive interference. In this case, ${\bf M}^{(c)}$, which is between $0$ and $1$, cannot bring the estimate close to ${\bf R}^{(c)}$. As a remedy, we consider an alternative objective, denoted as masked magnitude spectrum approximation (mMSA), which approximates ${\bf R}^{(c)}$ as the output of a masking operation on the mixture using a reference mask ${\bf O}^{(c)}$, such that ${\bf O}^{(c)} \odot  {\bf S}\approx {\bf R}^{(c)}$, for source $c$:
\begin{equation}
    \mathcal{L}_{\text{mMSA}} 
    = \sum_c ||({\bf O}^{(c)} - {\bf M}^{(c)}) \odot {\bf S}||_2^2.
\end{equation}
Note that this is equivalent to a weighted mask approximation objective, using the mixture magnitude as the weights. 

%
%
We can also define a global objective for the whole network as
\begin{equation}
    \mathcal{L}_{\text{CHI}} = \alpha \frac{\mathcal{L}_{\text{DC}}}{T F} + (1 - \alpha) \mathcal{L}_{\text{MI}}
\end{equation}
where $\alpha \in [0,1]$ controls the importance between the two objectives, and the objective $\mathcal{L}_{\text{MI}}$ for the mask inference head is either $\mathcal{L}_{\text{MSA}}$ or $\mathcal{L}_{\text{mMSA}}$. Note that here we divide $\mathcal{L}_{\text{DC}}$ by $T F$ because the objective for deep clustering calculates the pair-wise loss for each of the $(TF)^2$ pairs of T-F bins, while the spectrum approximation objective calculates end-to-end loss on the $TF$ time-frequency bins. 
For $\alpha=1$, only the deep clustering head gets trained together with the body, resulting in a deep clustering network. For $\alpha=0$, only the mask-inference head gets trained together with the body, resulting in a mask-inference network.

At test time, if both heads have been trained, either can be used. The mask-inference head directly outputs the T-F masks, while the deep clustering head outputs embeddings on which we perform clustering using, e.g., K-means.



\section{Evaluation and discussion}
\label{sec:typestyle}

\subsection{Datasets}
\label{sec:datasets}

For training and evaluation purposes, we built a remixed version of the DSD100 dataset for SiSEC \cite{DSD100}, which we refer to as DSD100-remix. For evaluation only, we also report results on two other datasets: the hidden iKala dataset for the MIREX submission, and the public iKala dataset for our newly proposed models. 


The DSD100 dataset includes synthesized mixtures and the corresponding original sources from 100 professionally produced and mixed songs. 
To build the training and validation sets of DSD100-remix, we use the DSD100 development set (50 songs). We design a simple energy-level-based detector \cite{ramirez2007voice} to remove silent parts in both the vocal and accompaniment tracks, so that the vocals and accompaniment fully overlap in the generated mixtures. After that, we downsample the tracks from 44.1~kHz to 16~kHz to reduce computational cost, and then randomly mix the vocals and accompaniment together at 0~dB SNR, creating a 15 h training set and a 0.5 h validation set.
We build the evaluation set of DSD100-remix from the DSD100 test set using a similar procedure, generating 50 pieces (one for each song) of fully-overlapped recordings with 30 seconds length each.


The input feature we use is calculated by the short-time Fourier transform (STFT) with 512-point window size and 128-point hop size. We use a 150-dimension mel-filterbank to reduce the input feature dimension. First-order delta of the mel-filterbank spectrogram is concatenated into the input feature. 
We used the ideal binary mask calculated on the mel-filterbank spectrogram as the target $Y$ matrix. 

\subsection{System architecture}


The Chimera network's body is comprised of 4 bi-directional long-short term memory (BLSTM) layers with 500 hidden units in each layer, followed by a linear fully-connected layer with a $D$-dimension vector output for each of the frame's $F=150$ T-F bins. Here, we use $D=20$ because it produced the best performance in a speech separation task~\cite{hershey2016deep}. In the mask-inference head, we set $C=2$ for the singing voice separation task, and use $\mathrm{softmax}$ as the non-linearity. 
We use the rmsprop algorithm \cite{tieleman2012lecture} as optimizer and select the network with the lowest loss on the validation set.


At test time, we split the signal into fixed-length segments, on which we run the network independently. We also tried running the network on the full input feature sequence, as in \cite{hershey2016deep}, but this lead to worse performance, probably due to the mismatch in context size between the training and test time. The mask-inference head of the network directly generates T-F masks. For deep clustering, the masks are obtained by applying K-means on the embeddings for the whole signal. 
We apply the mask for each source to the mel-filterbank spectrogram of the input, and recover the source using an inverse mel-filterbank transform and inverse-STFT with the mixture phase, followed by upsampling.



\subsection{Results for the MIREX submission}
\label{sec:results}

We first report on the system submitted to the Singing Voice Separation task of the Music Information Retrieval Evaluation eXchange (MIREX 2016) \cite{singing2016results}. That system only contains the deep clustering part, which corresponds to $\alpha=1$ in the hybrid system. In the MIREX system,  dropout layers with probability $0.2$ were added between each feed-forward connection, and sequence-wise batch normalization \cite{laurent2015batch} was applied in the input-to-hidden transformation in each BLSTM layer. Similarly to \cite{isik2016single}, we also applied a curriculum learning strategy \cite{bengio2009curriculum}, where we first train the network on segments of 100 frames, then train on segments of 500 frames. As distinguishing between vocals and accompaniment was part of the task, we used a crude rule-based approach: the mask whose total number of non-zero entries in the low frequency range ($<200$~Hz) is more than a half is used as the accompaniment mask, and the other as the vocals mask.

The hidden iKala dataset has been used as the evaluation dataset throughout MIREX 2014-2016, so we can report, as shown in Table~\ref{MIREX}, the results from the past three years, comparing the best two systems in each year's competition to our submitted system for 2016. The official MIREX results are reported in terms of global normalized SDR (GNSDR), global SIR (GSIR), global SAR (GSAR) \cite{singing2016mirex}.

Due to time limitations at the time of the MIREX submission, we submitted a system that we had trained using the DSD100-remix dataset described in Section~\ref{sec:datasets}. However, as mentioned in the MIREX description, the DSD100 dataset is different from both the hidden and public parts of the iKala dataset \cite{singing2016mirex}. Nonetheless, our system not only won the 1st place in MIREX 2016 but also outperformed the best systems from past years, even without training on the better-matched public iKala dataset, showing the efficacy of deep clustering for robust music separation. Note that the hidden iKala dataset is unavailable to the public, and it is thus unfortunately impossible to evaluate here what the performance of our system would be when trained on the public iKala data. 

\begin{table}[tbp]
\centering
\caption{Evaluation metrics for different systems in MIREX 2014-2016 on the hidden iKala dataset. \textit{V} denotes vocals and \textit{M} music.}\vspace{0.2cm}\label{MIREX}
\begin{tabular}{c|c|c|c|c|c|c}
\thline
 & \multicolumn{2}{c|}{GNSDR} & \multicolumn{2}{c|}{GSIR} & \multicolumn{2}{c}{GSAR}\\
 \hline
 & V & M & V & M & V & M \\
 \hline
\textbf{DC} &6.3&11.2&14.5&25.2&10.1&\phantom{1}7.3\\  
\hline
MC2 \cite{singing2016results}&5.3&\phantom{1}9.7&10.5&19.8&11.2&\phantom{1}6.1 \\
MC3 \cite{singing2016results}&5.5&\phantom{1}9.8&10.8&19.6&11.2&\phantom{1}6.3\\ 
FJ1 \cite{fan2016singing}&6.8&10.1&13.3&11.2&11.5&10.0\\ 
FJ2 \cite{fan2016singing}&6.3&\phantom{1}9.9&13.7&11.7&10.6&\phantom{1}9.1\\ 
IIY1 \cite{ikemiya2016singing}&4.2&\phantom{1}7.8&15.5&12.4&\phantom{1}7.7&\phantom{1}5.4\\
IIY2 \cite{ikemiya2016singing}&4.5&\phantom{1}7.9&13.3&14.3&\phantom{1}8.6&\phantom{1}5.0\\
\thline
\end{tabular}
\end{table}

\subsection{Results for the proposed hybrid system}
\label{sec:hybrid_results}


We now turn to the results using the Chimera networks. 
During the training phase, we use 100 frames of input features to form fixed duration segments. We train the Chimera network in three different regimes: a pure deep clustering regime ($\text{DC}$, $\alpha=1$), a pure mask-inference regime ($\text{MI}$, $\alpha=0$), and a hybrid regime  ($\text{CHI}_\alpha$, $0<\alpha<1$). All networks are trained from random initialization, and no training tricks mentioned above for the MIREX system are added. 
We report results on the DSD100-remix test set, which is matched to the training data, and the public iKala dataset, which is not.

By design, deep clustering provides one output for each source, and the sequence of the separation result is random. Therefore, the scores are computed  by using the best permutation between references and estimates at the file level.
Table \ref{tab:all} shows the results with the MSA objective in the MI head. 
We compute the source-to-distortion ratio (SDR), defined as scale-invariant SNR \cite{isik2016single}, for each test example, and report the length-weighted average over each test set of the improvement of SDR in the estimate with respect to that in the mixture (SDRi).

As can be seen in the results, $\text{MI}$ performs competitively with $\text{DC}$ on DSD100-remix, however DC  performs significantly better on the public iKala data. This shows the better generalization and robustness of the deep clustering method in cases where the test and training set are not matched. The best performance is achieved by $\text{CHI}_\alpha$-$\text{MI}$, the MI head of the Chimera network. Interestingly, the performance of the DC head does not change significantly for the values of $\alpha$ tested. This suggests that joint training with the deep clustering objective allows the body of the network to learn a more powerful representation than using the mask-inference objective alone; this representation is then best exploited by the mask-inference head thanks to its signal approximation objective. 

\begin{table}[tbp]
\centering
\caption{SDRi (dB) on the DSD100-remix and the public iKala datasets. The suffix after $\text{CHI}_\alpha$ denotes which head of the Chimera network is used for generating the masks. 
}\vspace{0.2cm}\label{tab:all}
\begin{tabular}{c|c|c|c|c}
\thline
 & \multicolumn{2}{c|}{DSD100-remix} & \multicolumn{2}{c}{iKala}\\
 \hline
 & \hspace{.4cm}V\hspace{.4cm} & \hspace{.4cm}M\hspace{.4cm} & \hspace{.4cm}V\hspace{.4cm} & \hspace{.4cm}M\hspace{.4cm}\phantom{}\\
 \thline
$\text{DC}$ &4.9&7.2& 6.1 & 10.0\\
\hline
$\text{MI}$ &4.8&6.7&5.2&\phantom{1}8.9 \\
\hline



$\text{CHI}_{0.1}$-DC &4.8&7.2&6.0&\phantom{1}9.7 \\
$\text{CHI}_{0.1}$-MI &\bf{5.5}&\bf{7.8}&\bf{6.4}&\bf{10.5} \\

$\text{CHI}_{0.5}$-DC &4.7&7.1&5.9&\phantom{1}9.9 \\
$\text{CHI}_{0.5}$-MI &\bf{5.5}&\bf{7.8}&6.3&\bf{10.5} \\
\thline
\end{tabular}
\end{table}

\begin{table}[tbp]
\centering
\caption{SDRi (dB) on the DSD100-remix and the public iKala datasets with various objectives in the MI head and embedding dimensions $D$.
}\vspace{0.2cm}\label{tab:obj}
\begin{tabular}{c|c|c|c|c|c}
\thline
\multicolumn{2}{c|}{} & \multicolumn{2}{c|}{DSD100-remix} & \multicolumn{2}{c}{iKala}\\
 \hline
$\mathcal{L}_{\text{MI}}$& $D$ & \hspace{.4cm}V\hspace{.4cm} & \hspace{.4cm}M\hspace{.4cm} & \hspace{.4cm}V\hspace{.4cm} & \hspace{.4cm}M\hspace{.4cm}\phantom{}\\
 \thline
$\text{\phantom{m}MSA}$ & 20 &\bf{5.5}&7.8&6.4&10.5 \\
$\text{mMSA}$& 20 & 5.4 & 7.8 & 6.5 & 10.7\\
$\text{mMSA}$& 10 & \bf{5.5} & \bf{7.9} & \bf{6.6} & \bf{10.8} \\
\thline
\end{tabular}
\end{table}

We now look at the influence of the objective used in the MI head. For the $\text{mMSA}$ objective, we use the Wiener like mask \cite{erdogan2015phase} since it is shown to have best performance among oracle masks computed from source magnitudes. As shown in Table~\ref{tab:obj}, training a hybrid $\text{CHI}_{0.1}$ network using the $\text{mMSA}$ objective leads to slightly better MI performance overall compared to $\text{MSA}$. 
We also consider varying the embedding dimension $D$, and find that reducing it from $D=20$ to $D=10$ leads to further improvements.
Because the output of the linear layer ${\bf Z}_t$ has dimension $F\times D$, decreasing $D$ also leaves room to increase the number of frequency bins $F$.

Table \ref{tab:feature} shows the results for various input features. We design various features by varying the sampling rate, the window/hop size in the STFT, and the dimension of the mel-frequency filterbanks. All networks are trained in the same hybrid regime as above, with the mMSA objective in the MI head and an embedding dimension $D=10$. For simplicity, we do not concatenate first-order deltas into the input feature. We can learn from the results that higher sampling rate, larger STFT window size STFT, and more mel-frequency bins result in better performance.

\begin{table}[tbp]
\centering
\caption{SDRi (dB) on the DSD100-remix and the public iKala datasets with various input features.
}\vspace{0.2cm}\label{tab:feature}
\begin{tabular}{c|c|c|c|c}
\thline
 & \multicolumn{2}{c|}{DSD100-remix} & \multicolumn{2}{c}{iKala}\\
 \hline
 & \hspace{.3cm}V\hspace{.3cm} & \hspace{.3cm}M\hspace{.3cm} & \hspace{.3cm}V\hspace{.3cm} & \hspace{.3cm}M\hspace{.3cm}\phantom{}\\
 \thline
$\text{16k-1024-256-mel150}$ & 5.5 & 7.9 & 6.6 & 10.6\\
 
$\text{16k-1024-256-mel200}$ & 5.5 & 7.9 & 6.9 & 10.9\\

$\text{22k-1024-256-mel200}$ & 5.9 & 7.9 & 7.2 & 10.7 \\

$\text{22k-2048-512-mel300}$ & \bf{6.1} & \bf{8.1} & \bf{7.4} & \bf{11.0} \\
\thline
\end{tabular}
\end{table}

\begin{figure}[ht]\vspace{0.0cm}
    \centering
    \includegraphics[width=\columnwidth]{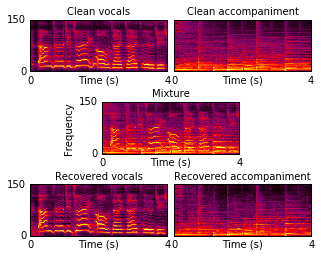}
    \caption{Example of separation results for a 4-second excerpt from file $\text{45378\_chorus}$ in the public iKala dataset.}
    \label{fig:spec}\vspace{0.0cm}
\end{figure}

\section{Conclusion}

In this paper, we investigated the effectiveness of a deep clustering model on the task of singing voice separation. Although deep clustering was originally designed for separating speech mixtures, we showed that this framework is also suitable for separating sources in music signals. Moreover, by jointly optimizing deep clustering with a classical mask-inference network, the new hybrid network outperformed both the plain deep clustering network and the mask-inference network. Experimental results confirmed the robustness of the hybrid approach in mismatched conditions.

\vspace{.2cm}
\noindent Audio examples are available at \cite{chimera_exp}.

\section{Acknowledgement}
The work of Yi Luo, Zhuo Chen, and Nima Mesgarani was funded by a grant from the National Institute of Health, NIDCD, DC014279, National Science Foundation CAREER Award, and the Pew Charitable Trusts. 


\vfill\pagebreak
\bibliographystyle{IEEEbib}
\bibliography{refs}

\end{document}